\documentclass[12pt]{iopart}

\expandafter\let\csname equation*\endcsname\relax
\expandafter\let\csname endequation*\endcsname\relax
\usepackage{amsmath} 
\usepackage{amssymb} 
\usepackage{hyperref}
\usepackage{graphicx} 
\usepackage{mathrsfs} 
\usepackage{fullpage}
\usepackage[usenames,dvipsnames]{color} 
\usepackage{ulem}

\usepackage{caption}
\usepackage{subcaption}

\usepackage{pdfpages}

\numberwithin{equation}{section}

\begin{document}

\title{Genetic optimization of the Hyperloop route through the Grapevine}
\author{Casey J. Handmer\\
  350-17, California Institute of Technology, Pasadena, California
  91125, USA}
\ead{chandmer@caltech.edu}

\begin{abstract}
We demonstrate a genetic algorithm that employs a versatile fitness function to optimize route selection for the Hyperloop, a proposed high speed passenger transportation system.
\end{abstract}

\section{The Hyperloop and the Grapevine: an awkward relationship}

The Hyperloop is a proposed rapid transportation system ideally suited to linking two metropolitan areas separated by between 200km and 1500km, such as Los Angeles and San Francisco~\cite{HyperloopWP}. Employing an elevated steel tube with reduced pressure, the Hyperloop is intended to reduce travel times between these two cities to around 30 minutes. 

\begin{figure}[h!]
  \centering
  \includegraphics[width=0.8\textwidth]{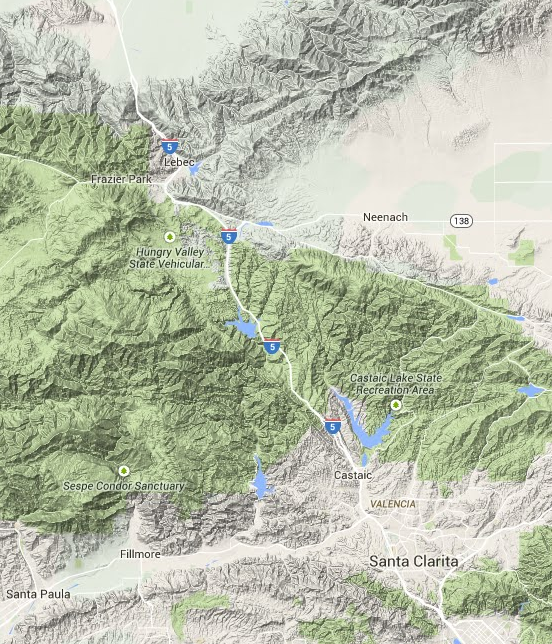}
  \caption{\small{Google maps screen grab of the Grapevine region, showing the I5 corridor and relief. The Hyperloop must traverse these mountains between lower right and upper left.}}
  \label{fig:Init}
\end{figure}

While the majority of the projected route follows the relatively straight Interstate 5 (I5) through the California central valley, entering the LA basin presents a more formidable challenge. Indeed, the existing rail transportation corridor must detour through the Tehachapi and Cajon passes. The I5 passes through a series of steep valleys between Santa Clarita and Tejon ranch, making several sharp turns. Ideally, the Hyperloop should follow a similar path, with bridges and tunnels smoothing the kinks enough to ensure passenger comfort even while travelling up to ten times faster than a car.

In practice, passenger discomfort can be avoided by making sufficiently gentle turns. The minimum radius of curvature is given by 
\begin{equation}
r_{min} = v^2/a_{max}\;.
\end{equation}
At a projected maximum speed of 1,220km/h (339m/s) and a maximum lateral acceleration of half a \(g\), \(r_{min} = 23.4\)km. Needless to say, the center of curvature must also smoothly vary---it wouldn't do to rattle passengers back and forth every second. The Hyperloop is proposed to traverse these hills and bends at a slower speed, but maximizing the minimum radius of curvature is still strongly favored.

The problem is thus defined. Bridges and tunnels are expensive. The region's geography is highly non-trivial. The route must employ gentle turns, ideally no less than 23.4km radius of curvature. This problem has an absurdly high dimensionality, and is thus well suited to the application of genetic algorithms: the second-best algorithm to solve any problem. 

\section{How to design a non-terrible genetic algorithm}

Genetic algorithms are not necessarily a cure-all. As anyone who has reproduced can tell you, genetics is a complicated business. The following elements are necessary, but not necessarily sufficient, to create a genetic algorithm that doesn't waste everyone's time~\cite{PiTPlecture}.

\subsection{Tight linkage}
Genetic algorithms use a genetic code, or genome, to parametrize a particular proposed solution. It is necessary to pick a representation that is sufficiently dense that phenotype (the physical nature of the proposal) is actually related to genotype (it's coded form). There is an art to it, and we found that a Bezier curve~\cite{Bezier}, parametrized by a set of order 10 control points, was ideal for this purpose. In essence, each genome contains a single gene.

\subsection{Useful diversity}
A genetic algorithm that introduces too much mutation in each generation is equivalent to a random search, and thus hopelessly inefficient. Conversely, too little or too limited mutation will result in premature stagnation with no exploration of the solution space. Some algorithms can dynamically vary mutation parameters. Similarly, an algorithm that begins with, say, 100 different genomes will quickly eliminate the weaker ones, narrowing an already limited gene pool. To avoid stagnation due to inbreeding, one can introduce a small number of new, randomly generated genomes at every generation.

\subsection{Good variation operators}
In addition to tweaking mutation parameters and ensuring a steady flow of new genetic material into the system, genetic algorithms must also perform crossover. Sometimes known as ``the fun part'', crossover occurs where two selected genomes mix their parameters to create the next generation. Mutation is typically introduced at this stage.

\subsection{A good fitness function}
This is the hardest part. A well designed genetic algorithm will find an optimal solution to the stated problem. If the problem is poorly stated, through poor design of a fitness function, then the solution will be unsatisfying. In the Hyperloop problem, there are several metrics to optimize, including cost, curvature, and maximum grade. Constructing a multidimensional Pareto front is preferable to an arbitrarily chosen weighted average. In more complicated problems, co-evolving fitness criteria is necessary to find a suitable global solution. In this instance, comparing the genome fitness with randomly generated sequences is a good way to normalize between generations.

\section{Technical implementation}
We implemented the Hyperloop routing genetic algorithm in {\it Mathematica}, included in Appendix~A. For the purposes of routing, we chose to start at the I5-I405 interchange near Granada Hills (GPS:34.29,-118.47) and end at the Tesla Supercharger in Tejon Ranch (GPS:34.99,-118.95).

\subsection{Geodata}
Geodata was obtained from the USGS website~\cite{USGS}, corresponding to the n35w119 graticule with 1 arc-second resolution in ArcGrid format.

\subsection{Fitness function}
Our fitness function assumed a constant cost per length of tunnel, and a cost for bridges or pylons that scaled with both height and length. Mathematically expressed, cost in \$/m is
\begin{align}
cost_{pylon} &= 116 h_{pylon}^2 \;,\\
cost_{tunnel} &= 310000 \;.
\end{align}
For a typical pylon height of 6m, this gives a cost per km of \$4.2m. For tunnels, the whitepaper's suggestion of \$31m/km is probably an order of magnitude too low as the Hyperloop will require at least two vacuum tunnels and a service tunnel~\cite{tunnelcost}. Additionally, the route has to cross the San Gabriel,  San Andreas, and Garlock faults, increasing geological complexity. We used a cost of \$310m/km. It is worth noting that placement of an optimal corridor is not highly sensitive to cost-function parameters.

The constraint on radius of curvature and maximum grade (6\%) was applied by calculating both using finite differences and then applying a steep penalty cost if limits were exceeded.

\subsection{Initialization}

The population is initialized with a specified number (we used 200) of genomes. Genomes are generated by randomly generating control points within the space. A specialized function converts them to a physical path from which metrics can be calculated. One such initial condition is shown in Fig.~\ref{fig:Init}. This population is entirely suboptimal, with paths that do not remotely follow the landscape, involve sharp turns, and pylons or bridges that vastly exceed today's engineering capability. Within a hundred generations of the algorithm, however, the evolved population will be much better suited to the problem landscape.

\begin{figure}[h!]
  \centering
  \includegraphics[width=0.8\textwidth]{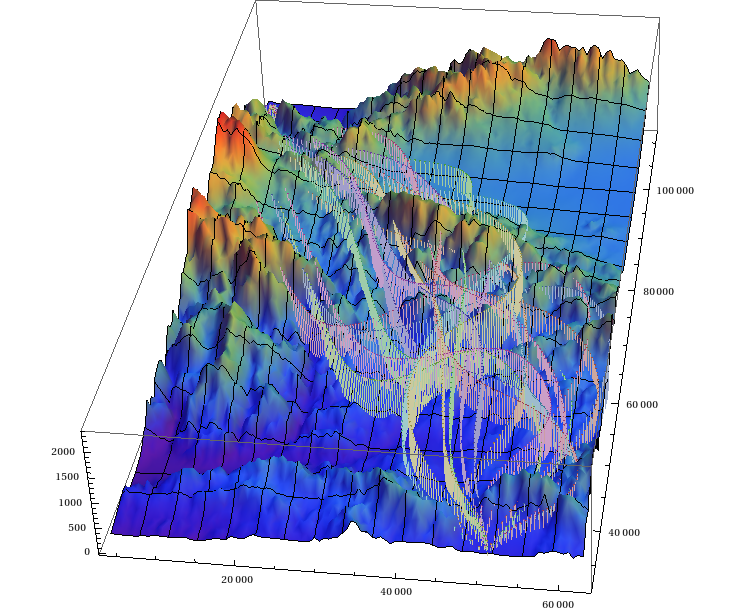}
  \caption{\small{Relief map showing a subset of the initial population superposed on the relevant geography. The vertical scale is greatly exaggerated, distances are in meters.}}
  \label{fig:Init}
\end{figure}

\section{Results}

We found two solution classes that were locally optimal. The first underlines the challenges of using genetic algorithms, wherein a simulation found a simple straight tunnel 300m above sea level through the entire range.

\begin{figure}[h!]
  \centering
  \includegraphics[width=0.8\textwidth]{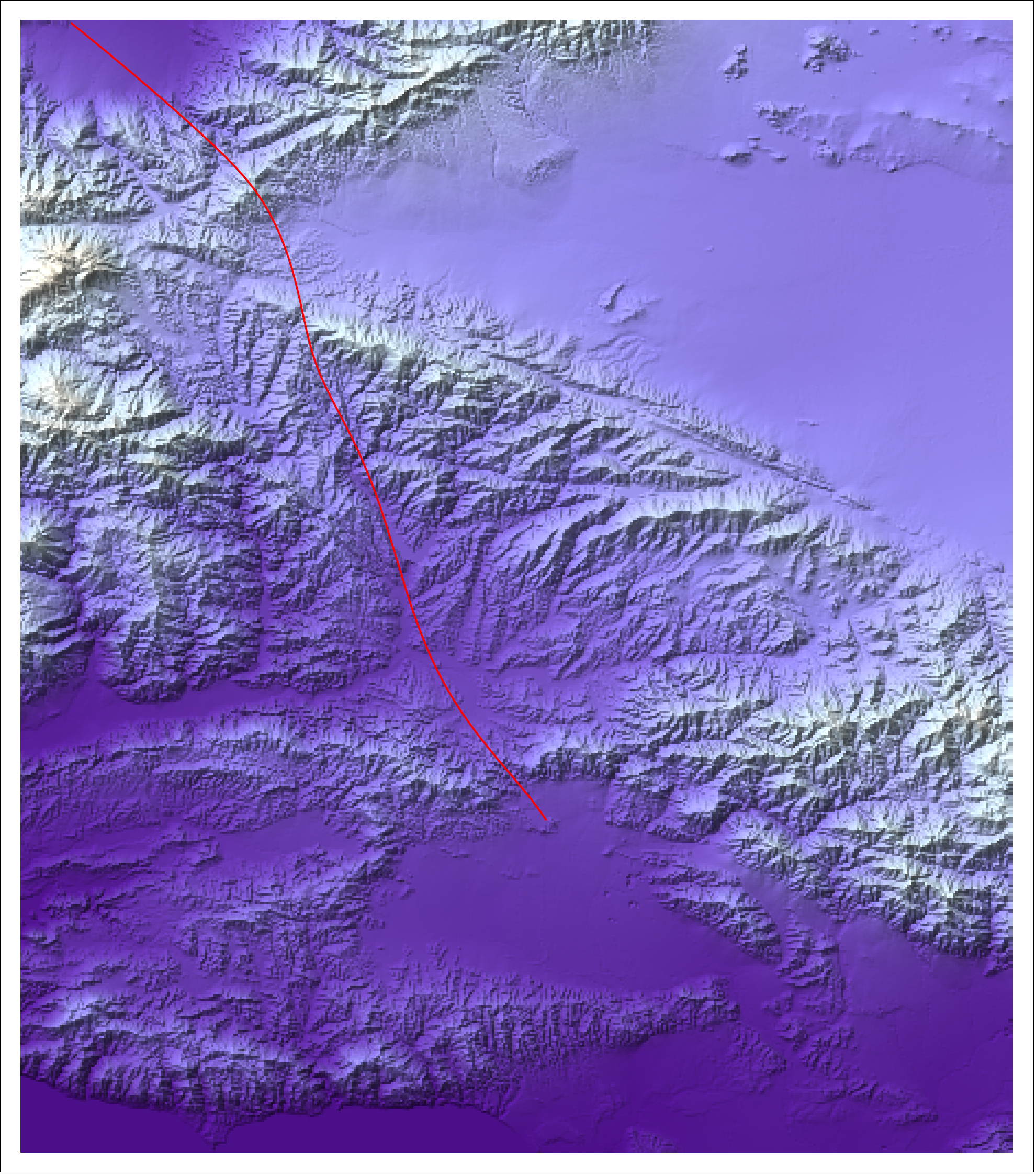}
  \caption{\small{Diagram showing proposed route on a relief map.}}
  \label{fig:Relief}
\end{figure}

\begin{figure}[h!]
  \centering
  \includegraphics[width=0.8\textwidth]{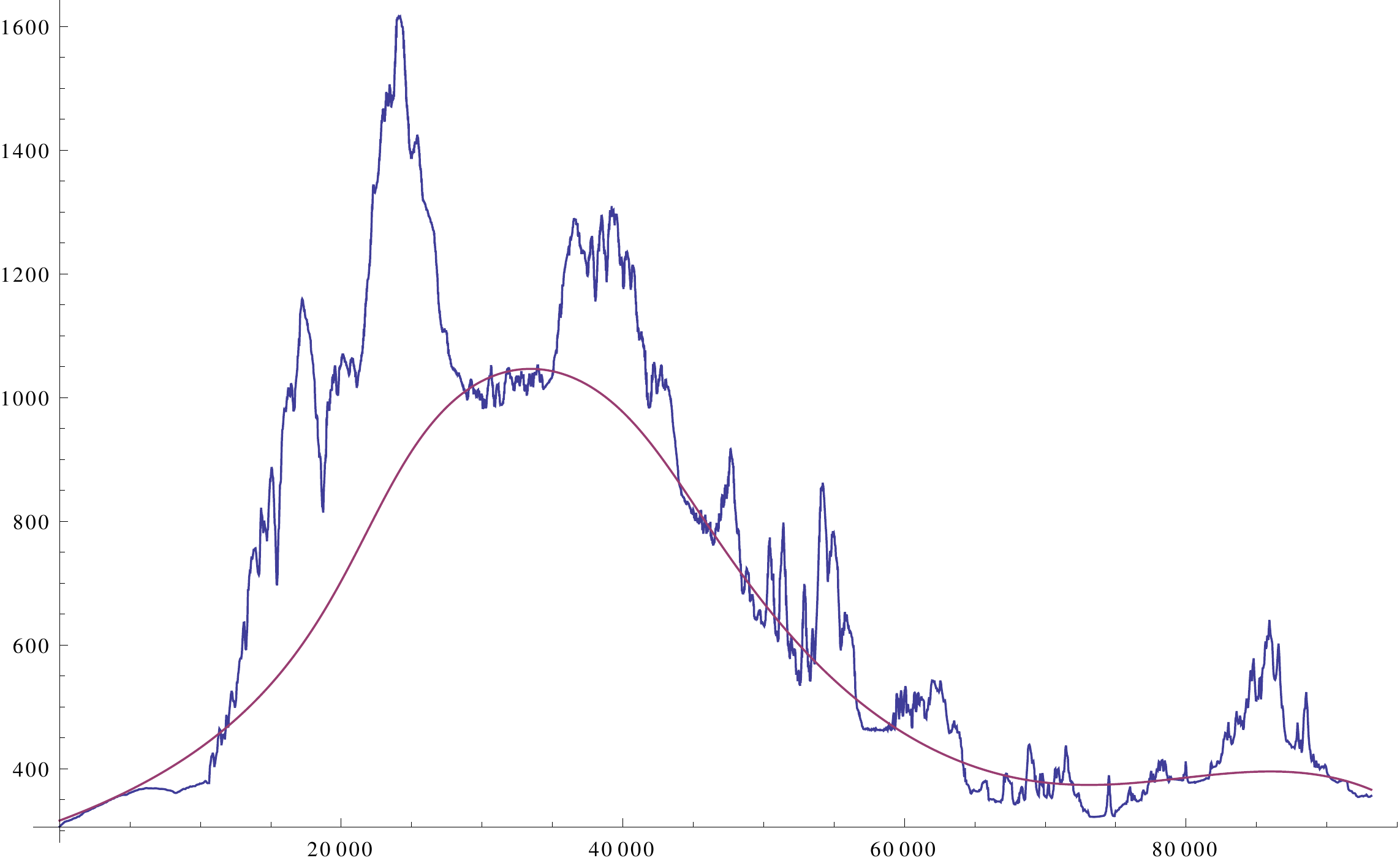}
  \caption{\small{Diagram showing profile of route, with implied bridges and tunnels. Distances in meters, vertical scale exaggerated by a factor of 73.}}
  \label{fig:Profile}
\end{figure}

The other, much more interesting solution set initially follows the I5 corridor north, before diverting to the east over Castaic Lake, cutting across the western extremity of Antelope Valley and finally returning to the surface of the central valley. This solution is shown in Figs.~\ref{fig:Relief} and~\ref{fig:Profile}.

It features a minimum curvature radius of 20km, a mean pylon height of 22m, a total tunnel length of 48km, and a maximum subsurface depth of 738m. A nearby variant shown in Appendix A achieves a minimum curvature radius of 23.5km for the use of 54km of tunnels---enabling transit through the Grapevine without slowing until the destination.

\section{Conclusion}
We have applied a genetic algorithm to solve a high dimensionality optimization problem. We have found a route for the Hyperloop through the Grapevine that permits a transit speed of 1130km/h (314m/s), reducing transit time by 139 seconds compared to that proposed in the whitepaper. More importantly, we have demonstrated the viability of an evolutionary approach to high speed transport route finding in general.

\section*{Appendix A}



\section*{Supplementary material}

# Hyperloop genetic algorithm code

## Path

```
path = "/home/user/Downloads/";
```

## Import data!

```
t = Import[path <> "n35w119.zip", "Data"];
```

In[3]:= `Dimensions[t]`

Out[3]= {3612, 3612}

In[5]:= `ReliefPlot[t[[1 ;; -1 ;; 10, 1 ;; -1 ;; 10]]]`

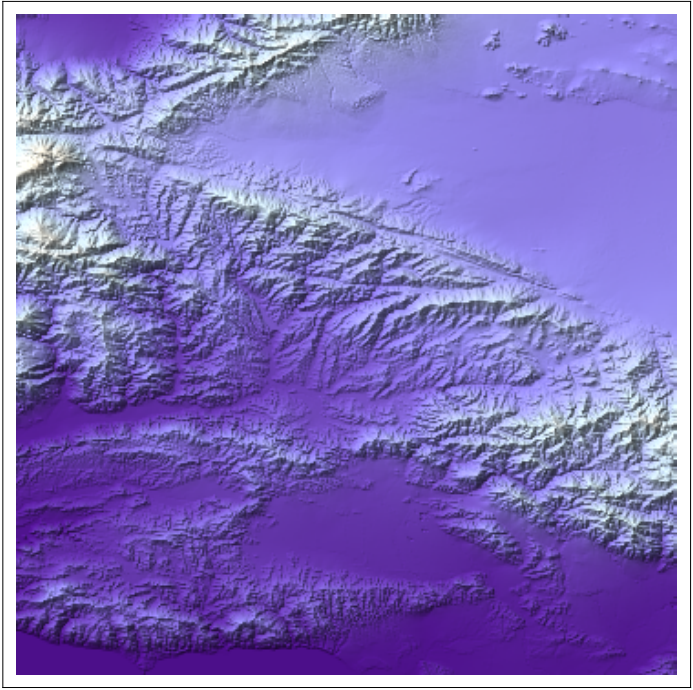

Out[5]=

## x (east) and y (north) measured in meters. Origin bottom left. Posmults convert arc seconds to meters in each direction.

In[8]:= `posmults = {26.729179129149337`, 30.864197530864196`};`

### Perform basic interpolation on data for a fine-grained elevation function.

```
Elevation[x_, y_] := Catch[
   If[Floor[y / posmults[[2]]] < 2 || Floor[y / posmults[[2]]] > 3611 ||
      Floor[x / posmults[[1]]] < 2 || Floor[x / posmults[[1]]] > 3611, Throw[10.^10];];
   Clear[dat]; dat = Flatten[Table[{i, j, t[[Floor[y / posmults[[2]]] + i,
          Floor[x / posmults[[1]]] + j]]}, {i, -1, 1}, {j, -1, 1}], 1];
   Throw[Fit[dat, {1, a, b, a^2, a b, b^2}, {a, b}] /.
     {a → Mod[x / posmults[[1]], 1], b → Mod[y / posmults[[2]], 1]}]]
```

### Define endpoints of path. uleft is Tejon Ranch Supercharger. lright is the 405-5 interchange.

```
In[135]:= uleft = Reverse[Mod[{34.996816, -118.948658}, 1]] Length[t];
         lright = Reverse[Mod[{34.293672, -118.470066}, 1]] Length[t];
```

### Baseline - a straight tunnel from end to end.

```
In[139]:= trackstraight =
           Table[Join[uleft posmults + (lright posmults - uleft posmults) * i, {300}],
            {i, 0, 1, 1.0 / 3200}];
```

### Compute lengths of sides of a triangle specified by three points

```
In[12]:= SideLengths[{point1_, point2_, point3_}] :=
           {Sqrt[(point1 - point2).(point1 - point2)], Sqrt[
             (point3 - point2).(point3 - point2)], Sqrt[(point1 - point3).(point1 - point3)]};
```

### Compute the implied radius of curvature of any triangle given side lengths

```
In[13]:= Radius[{a_, b_, c_}] := If[a + b ⩵ c || a + c ⩵ b || b + c ⩵ a,
           ∞, a b c / Sqrt[(a + b + c) (-a + b + c) (a - b + c) (a + b - c)]];
```

### Compute the tightest curve in any given track

```
In[14]:= MinCurv[track_] :=
           Min[Table[Radius[SideLengths[track[[i ;; i + 2]]]], {i, 1, Length[track] - 2}]];
```

### Compute maximum depth of tunnels - useful to diagnose subterranean meandering

```
In[15]:= ElevCheck[track_] := Min[Table[
            track[[i, 3]] - Elevation[track[[i, 1]], track[[i, 2]]], {i, 1, Length[track]}]];
```

### BezFunc converts a genome (points) to an affine function. It adds

control points for the start and end points, assuming 10m elevation.

```
In[20]:= BezFunc[points_] := BezierFunction[
    Join[{Join[uleft posmults, {Apply[Elevation, uleft posmults] + 10}]}, points,
     {Join[lright posmults, {Apply[Elevation, lright posmults] + 10}]}]];
```

TrackBez converts a genome (points) to 3701 points evenly spaced along the proposed path.

```
In[21]:= TrackBez[points_] := Catch[
    Clear[fn];
    fn = BezFunc[points];
    Throw[Table[fn[t], {t, 0, 1, 1 / 3700}]]];
```

The fitness (cost) function. Computes the expense of a proposed path parametrized by the genome points. A lot of functions are manually recomputed here to save time, because *Mathematica*.

```
Cost[points_] := Catch[
   Clear[mincurv, tr, spacing, grade,
    spacing3, spacing32, stepdif, curvature, pyheight];
   mincurv = {12 000, 16 000, 23 500}[[2]]; (*Set the floor on curvature*)
   tr = DeleteDuplicates[TrackBez[points]]; (*Define the track*)
spacing =
    Sqrt[Apply[Plus, Transpose[(tr[[2 ;; -1]] - tr[[1 ;; -2]])[[All, 1 ;; 2]]^2]]];
   (*compute the spaces between each point in tr*)
grade = (tr[[2 ;; -1]] - tr[[1 ;; -2]])[[All, 3]] / spacing;
   (*compute the gradient at every point*)
spacing3 = Sqrt[Apply[Plus, Transpose[(tr[[2 ;; -1]] - tr[[1 ;; -2]])^2]]];
   (*compute triangle side lengths*)
spacing32 = Sqrt[Apply[Plus, Transpose[(tr[[3 ;; -1]] - tr[[1 ;; -3]])^2]]];
   (*compute more triangle side lengths*)
stepdif = Transpose[{spacing3[[1 ;; -2]], spacing3[[2 ;; -1]], spacing32}];
   (*integrate triangle data*)
curvature = Apply[Times, Transpose[stepdif]] /
     Sqrt[stepdif.{1, 1, 1} stepdif.{-1, 1, 1} stepdif.{1, -1, 1} stepdif.{1, 1, -1}];
   (*compute curvature at each point*)
pyheight = tr[[All, 3]] - Table[Elevation[tr[[i, 1]], tr[[i, 2]]], {i, 1, Length[tr]}];
   (*compute height or depth of track relative to ground level*)
   (*Apply penalties for excessive curvature or grade as a multiplicative
    factor to the raw cost function, computed from pyheight*)
Throw[(0.005 (-Min[curvature] + mincurv + Abs[Min[curvature] - mincurv]) + 1)
     (100 (Max[grade] - 0.06 + Abs[Max[grade] - 0.06]) + 1)
     Total[spacing ((0.5 (Abs[pyheight] + pyheight))[[1 ;; -2]]^2 * 116.0 + 310 000.0
          ((-Abs[pyheight + 10] + pyheight + 10) / (2.0 pyheight + 20))[[1 ;; -2]])]]];
```

## Compute average pylon height

```
In[23]:= AveragePyHeight[points_] := Catch[
    Clear[tr, pyheight];
    tr = TrackBez[points];
    pyheight =
     tr[[All, 3]] - Table[Elevation[tr[[i, 1]], tr[[i, 2]]], {i, 1, Length[tr]}];
    Throw[Total[0.5 (Abs[pyheight] + pyheight)] / Total[0.5 Sign[pyheight] + 0.5]]];
```

## Compute length of tunnels, where the track is deeper than 10m below surface (above which excavation will suffice)

```
In[121]:= TunnelLength[points_] := Catch[
    Clear[tr, pyheight, spacing];
    tr = TrackBez[points];
    spacing =
     Sqrt[Apply[Plus, Transpose[(tr[[2 ;; -1]] - tr[[1 ;; -2]])[[All, 1 ;; 2]]^2]]];
    pyheight = tr[[All, 3]] - Table[Elevation[tr[[i, 1]], tr[[i, 2]]] - 10,
       {i, 1, Length[tr]}];
    Throw[(-0.5 Sign[pyheight] + 0.5)[[2 ;; -1]].spacing]];
```

## Genetic algorithm

### Genetic algorithm functions

```
Crossover[points1_, points2_] := Catch[(*Generating new individuals*)
    Clear[rnd]; rnd = Round[RandomReal[{0, 1}, Dimensions[points1]]];
    Throw[rnd * points1 + (1 - rnd) * points2 +
      Round[RandomReal[{0, 0.8}]] RandomReal[{-1, 1}, Dimensions[points1]]
       100 RandomReal[{0, 1}] Table[{100, 100, 1}, {Length[points1]}]]];

Rank[pop_] := Sort[Table[{Cost[pop[[i]]], i}, {i, 1, Length[pop]}]];
(*ranking individuals*)

Breed[subpop_] := Catch[(*arranging marriages*)
    Clear[mothers, fathers];
    mothers = RandomInteger[{1, Length[subpop]}, Length[subpop]];
    fathers = RandomInteger[{1, Length[subpop]}, Length[subpop]];
    Throw[Join[subpop, Table[Crossover[subpop[[mothers[[i]]]],
        subpop[[fathers[[i]]]]], {i, 1, Length[subpop]}]]]];

In[234]:= Generation[pop_] := Catch[(*putting it all together*)
    Clear[parentsi, parents];
    parentsi = Rank[pop][[1 ;; Length[pop] / 2, 2]];
    parents = Table[pop[[parentsi[[i]]]], {i, 1, Length[parentsi]}];
    Throw[Join[Breed[parents][[1 ;; -Round[0.02 Length[population]] - 1]],
      NewPopulation[Round[0.02 Length[population]]]]];
```

## Randomly generate new candidates

```
In[162]:= NewPopulation[n_] :=
    3600 RandomReal[{0, 1}, {n, 8, 3}] Table[Table[Join[posmults, {0.5}], {8}], {n}];
```

```
In[349]:= population = NewPopulation[200]; (*Randomly generated initial population*)
```

## Close to optimal solution

```
In[341]:= (*314m/s, 48km of tunnel*)
    champion = {{35435.85845872003`, 86397.78479923689`, 649.1296937001766`},
       {7192.091453202574`, 102932.26407319008`, 1190.9381775414531`},
       {42005.279175087046`, 92686.66501036017`, 1430.977185882561`},
       {33493.98753860159`, 59124.683268222885`, 731.7341099973344`},
       {-988.5600949717715`, 84198.89301272736`, 1226.9034240190824`},
       {64421.45257537803`, 81295.87604793729`, 476.3238278301031`},
       {20149.554097645694`, 52287.03145794095`, 9.464254857920556`},
       {41455.4673191837`, 47525.82103174714`, 534.6276877397107`}};
```

```
In[331]:= ProfilePlot[TrackBez[champion]]
```

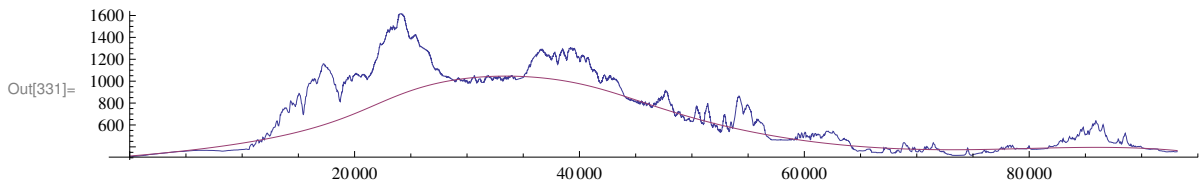

```
In[342]:= Cost[champion]
    ElevCheck[TrackBez[champion]]
    MinCurv[TrackBez[champion]]
    AveragePyHeight[champion]
    TunnelLength[champion]
```

Out[342]= $1.83587 \times 10^{10}$

Out[343]= $-738.196$

Out[344]= 20 093.3

Out[345]= 21.883

Out[346]= 47 670.6

```
(*342m/s, 54km of tunnels*)
champion = {{30367.446073026338`, 86786.1254069737`, 618.1502440459486`},
   {6317.24686505232`, 98748.85892332364`, 1113.3097545428625`},
   {38234.73635589907`, 86342.4199902501`, 1580.8125294686706`},
   {44172.94595445811`, 62595.533010094085`, 824.1037439325853`},
   {270.3823980015038`, 83642.07463046617`, 1096.9250673589777`},
   {55747.249044611155`, 78792.8285025906`, 467.50523242502106`},
   {19376.24986857262`, 50853.49116525148`, -61.808796983233115`},
   {42290.549458999085`, 52900.55410823617`, 553.6322528110212`}};
```

```
In[328]:= ProfilePlot[TrackBez[champion]]
```

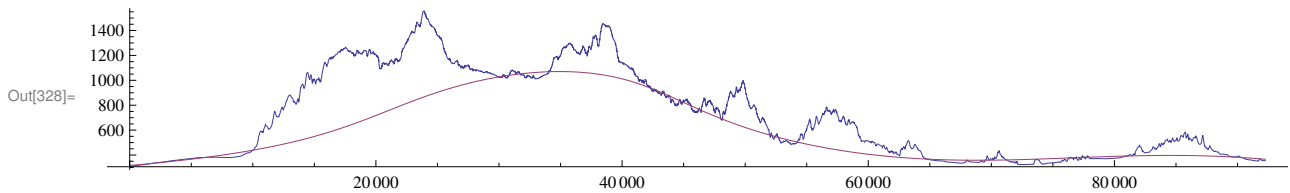

```
In[336]:= Cost[champion]
        ElevCheck[TrackBez[champion]]
        MinCurv[TrackBez[champion]]
        AveragePyHeight[champion]
        TunnelLength[champion]
```

Out[336]= $1.81901 \times 10^{10}$

Out[337]= $-688.886$

Out[338]= $23853.4$

Out[339]= $19.2651$

Out[340]= $53565.1$

```
In[295]:= champion = population[[1]];

        population = Table[champion RandomReal[{0.8, 1/0.8}, Dimensions[champion]], {100}];
        (*This allows kickstarting around a known successful solution*)
```

## Evaluate cost of initial population

```
In[311]:= Rank[population][[1 ;; 10]]
```

Out[311]= $\{\{2.03224 \times 10^{10}, 3\}, \{2.08142 \times 10^{10}, 77\}, \{2.09168 \times 10^{10}, 29\},$
$\{2.1302 \times 10^{10}, 59\}, \{2.13563 \times 10^{10}, 80\}, \{2.14179 \times 10^{10}, 74\},$
$\{2.17517 \times 10^{10}, 54\}, \{2.1953 \times 10^{10}, 96\}, \{2.19667 \times 10^{10}, 11\}, \{2.19799 \times 10^{10}, 23\}\}$

```
In[312]:= Cost[champion]
```

Out[312]= $1.81901 \times 10^{10}$

## Perform evolution for 200 generations, evaluate top performers

```
        Do[population = Generation[population], {200}]
        Rank[population][[1 ;; 10]]
```

Out[314]= $\{\{1.82877 \times 10^{10}, 79\}, \{1.83127 \times 10^{10}, 1\}, \{1.83684 \times 10^{10}, 2\},$
$\{1.839 \times 10^{10}, 68\}, \{1.8419 \times 10^{10}, 3\}, \{1.84442 \times 10^{10}, 4\}, \{1.84475 \times 10^{10}, 5\},$
$\{1.84518 \times 10^{10}, 93\}, \{1.84633 \times 10^{10}, 6\}, \{1.84672 \times 10^{10}, 7\}\}$

```
In[315]:= Cost[population[[1]]]
       ElevCheck[TrackBez[population[[1]]]]
       MinCurv[TrackBez[population[[1]]]]
       AveragePyHeight[population[[1]]]
       TunnelLength[population[[1]]]

Out[315]= 1.83127 × 10^10

Out[316]= -720.408

Out[317]= 20 615.2

Out[318]= 20.0805

Out[319]= 54 352.8
```

## Output data

```
In[16]:= PathPlot[track_] := Show[ListPointPlot3D[track[[1 ;; -1 ;; 20]], Filling → Bottom,
          FillingStyle → Red, PlotStyle → Directive[Red, PointSize[Small]],
          BoxRatios → Join[Abs[(lright - uleft)] posmults, {15 000}]],
         ListPlot3D[Flatten[Table[{x, y, Elevation[x, y]}, {x, Min[track[[All, 1]]],
             Max[track[[All, 1]]], (Max[track[[All, 1]]] - Min[track[[All, 1]]]) / 100},
            {y, Min[track[[All, 2]]], Max[track[[All, 2]]],
             (Max[track[[All, 2]]] - Min[track[[All, 2]]]) / 100}], 1], ColorFunction →
           "Rainbow", BoxRatios → {(Max[track[[All, 1]]] - Min[track[[All, 1]]]),
            (Max[track[[All, 2]]] - Min[track[[All, 2]]]), 15 000}]]

In[17]:= PathPopPlot[track_] :=
        Show[ListPointPlot3D[Table[track[[i]][[1 ;; -1 ;; 20]], {i, 1, Length[track]}],
          Filling → Bottom, PlotStyle → Directive[PointSize[Small]],
          BoxRatios → Join[Abs[(lright - uleft)] posmults, {15 000}]],
         ListPlot3D[Flatten[Table[{x, y, Elevation[x, y]},
            {x, Min[track[[All, All, 1]]], Max[track[[All, All, 1]]],
             (Max[track[[All, All, 1]]] - Min[track[[All, All, 1]]]) / 100},
            {y, Min[track[[All, All, 2]]], Max[track[[All, All, 2]]],
             (Max[track[[All, All, 2]]] - Min[track[[All, All, 2]]]) / 100}], 1],
          ColorFunction → "Rainbow", BoxRatios →
           {(Max[track[[All, All, 1]]] - Min[track[[All, All, 1]]]),
            (Max[track[[All, All, 2]]] - Min[track[[All, All, 2]]]), 15 000}]]

In[327]:= ProfilePlot[track_] := Catch[
         footlength = Table[0, {Length[track]}];
         Do[footlength[[i]] =
           footlength[[i - 1]] + Sqrt[(track[[i, 1 ;; 2]] - track[[i - 1, 1 ;; 2]]).
              (track[[i, 1 ;; 2]] - track[[i - 1, 1 ;; 2]])];, {i, 2, Length[track]}];
         Throw[ListLinePlot[{Table[{footlength[[i]], Apply[Elevation,
               track[[i, 1 ;; 2]]]}, {i, 1, Length[track]}],
            Transpose[{footlength, track[[All, 3]]}]}, AspectRatio → 0.137]]];
```

## Typical initial condition

```
In[350]:= Show[ReliefPlot[t[[1 ;; -1 ;; 7, 1 ;; -1 ;; 8]]],
     Table[ListLinePlot[# / Table[{8, 7} posmults, {Length[#]}] &[
        TrackBez[population[[i]]][[All, 1 ;; 2]]], PlotStyle → Red], {i, 1, 20}]]
```

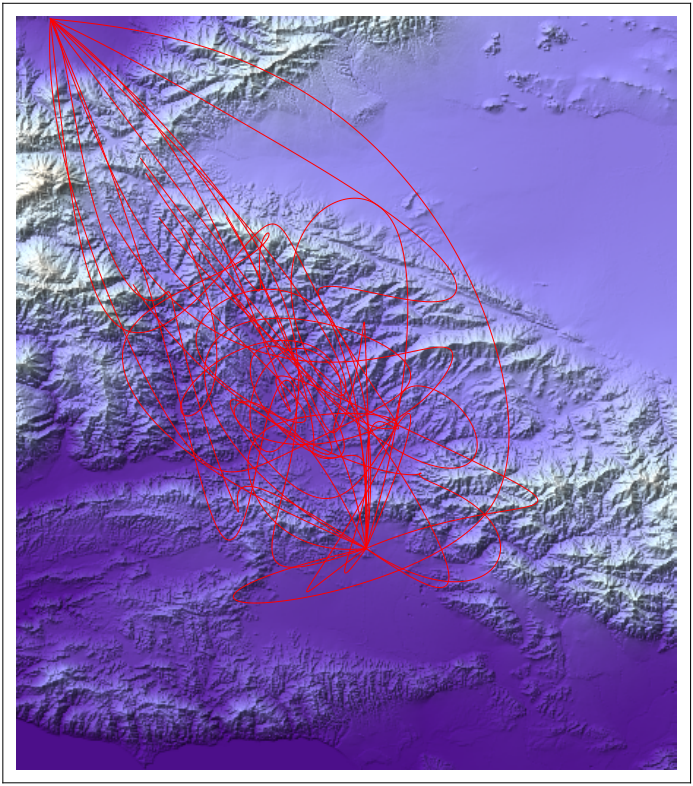

```
PathPopPlot[Table[TrackBez[population[[i]]], {i, 1, Min[20, Length[population]]}]]
(*Output (Figure 2) removed because it's 36Mb.*)
```

## Converged solution

```
In[348]:= Show[ReliefPlot[t[[1 ;; -1 ;; 7, 1 ;; -1 ;; 8]]],
    Table[ListLinePlot[# / Table[{8, 7} posmults, {Length[#]}] &[
        TrackBez[population[[i]]][[All, 1 ;; 2]]], PlotStyle → Red], {i, 1, 20}]]
```

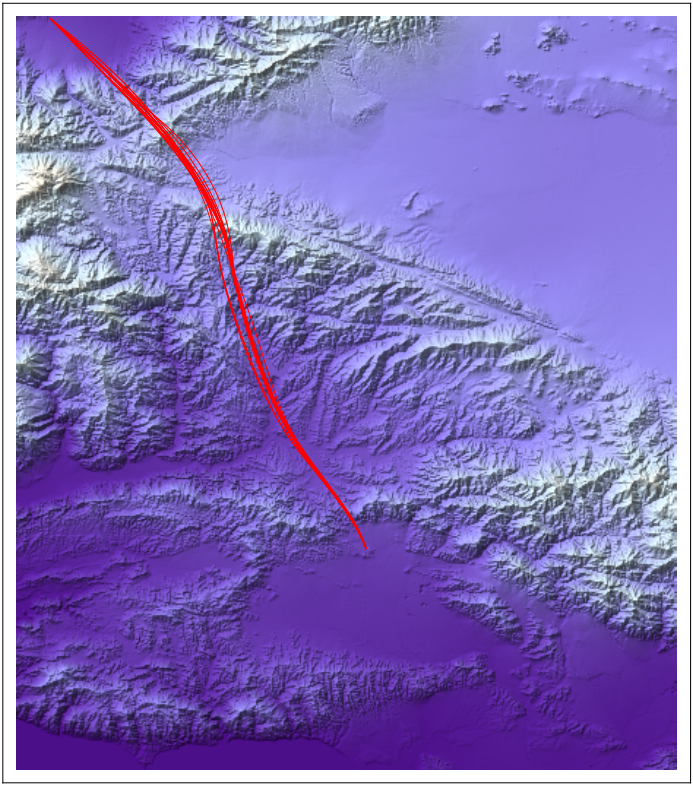




\section*{References}

\begin{thebibliography}{9}

\bibitem{HyperloopWP}
  \url{http://www.spacex.com/sites/spacex/files/hyperloop_alpha-20130812.pdf} 
  Accessed March 2 2015.

\bibitem{PiTPlecture}
  Lipson, Hod. ``Co-Evolutionary Learning: Distilling Free-Form Natural Laws from Experimental Data.'' Prospects in Theoretical Physics. IAS, Princeton, NJ. 20 July 2012. Lecture.

\bibitem{Bezier}
  \url{http://en.wikipedia.org/wiki/Bezier_curve}
  Accessed March 2 2015.

\bibitem{USGS}
  USGS data is available at the National Map Viewer.
  \url{http://viewer.nationalmap.gov}

\bibitem{tunnelcost}
  \url{https://pedestrianobservations.wordpress.com/2011/05/16/us-rail-construction-costs/} 
  Accessed March 2 2015.

\end{thebibliography}
\end{document}